\documentclass[
]{ceurart}

\usepackage{listings}
\usepackage{xcolor}
\usepackage{fancyvrb}
\usepackage{caption}
\usepackage{float}
\usepackage{tcolorbox}
\usepackage{booktabs}
\usepackage{multirow}
\usepackage{graphicx}
\usepackage{hhline}

\definecolor{myblue}{RGB}{0, 0, 205}
\definecolor{mygreen}{RGB}{0, 128, 0}

\begin{document}

\copyrightyear{2022}
\copyrightclause{Copyright for this paper by its authors.
  Use permitted under Creative Commons License Attribution 4.0
  International (CC BY 4.0).}

\conference{SemTab2024}

\title{Column Vocabulary Association (CVA): \\
semantic interpretation of dataless tables}

\author[]{Margherita Martorana}[%
orcid=0000-0001-8004-0464,
email=m.martorana@vu.nl,
]
\cormark[1]

\author[]{Xueli Pan}[%
orcid=0000-0002-3736-7047,
email=x.pan2@vu.nl,
]

\author[]{Benno Kruit}[%
email=b.b.kruit@vu.nl,
]

\author[]{Tobias Kuhn}[%
orcid=0000-0002-1267-0234,
email=t.kuhn@vu.nl,
]

\author[]{Jacco van Ossenbruggen}[%
orcid=0000-0002-7748-4715,
email=j.r.van.ossenbruggen@vu.nl,
]

\address[]{Department of Computer Science, Vrije Universiteit Amsterdam, De Boelelaan 1105, Amsterdam, The Netherlands}

\cortext[1]{Corresponding author.}

\begin{abstract}
Traditional Semantic Table Interpretation (STI) methods rely primarily on the underlying table data to create semantic annotations. This year's SemTab challenge introduced the ``Metadata to KG'' track, which focuses on performing STI by using only metadata information, without access to the underlying data. In response to this new challenge, we introduce a new term: Column Vocabulary Association (CVA). This term refers to the task of semantic annotation of column headers solely based on metadata information. In this study, we evaluate the performance of various methods in executing the CVA task, including a Large Language Models (LLMs) and Retrieval Augmented Generation (RAG) approach, as well as a more traditional similarity approach with SemanticBERT. Our methodology uses a zero-shot setting, with no pretraining or examples passed to the Large Language Models (LLMs), as we aim to avoid a domain-specific setting.

We investigate a total of 7 different LLMs, of which three commercial GPT models (i.e. \texttt{gpt-3.5-turbo-0.125}, \texttt{gpt-4o} and \texttt{gpt-4-turbo}) and four open source models (i.e. \texttt{llama3-80b}, \texttt{llama3-7b}, \texttt{gemma-7b} and \texttt{mixtral-8x7b}). We integrate this models with RAG systems, and we explore how variations in temperature settings affect performances. Moreover, we continue our investigation by performing the CVA task utilizing SemanticBERT, analyzing how various metadata information influence its performance.

Initial findings indicate that LLMs generally perform well at temperatures below 1.0, achieving an accuracy of 100\% in certain cases. Nevertheless, our investigation also reveal that the nature of the data significantly influences CVA task outcomes. In fact, in cases where the input data and glossary are related (for example by being created by the same organizations) traditional methods appear to surpass the performance of LLMs. 

\end{abstract}


\maketitle

\section{Introduction}

Tabular data is the most common format used for data storage and sharing\cite{shwartz2022tabular}. However, tabular data often lacks semantic annotations and can contain inaccurate or missing information. Semantic Table Interpretation (STI) aims to find semantic annotations for table cells and columns, as well as column relationships, using existing Knowledge Graphs (KGs). Semantic annotations are particularly important when used to enrich and augment metadata. In fact, several studies\cite{boeckhout2018fair,mons2018data,lamprecht2020towards} have shown that high-quality metadata supports data Findability, Accessibility, Interoperability and Reusability (FAIR Guiding Principles)\cite{wilkinson2016fair}. Rich metadata plays a critical role when dealing with confidential data, as the underlying data is not commonly openly and freely accessible. Enhancing the FAIRness for this type of data has gained more attention in recent years, and previous work has suggested that high-quality and rich metadata improves the discovery and reuse of these resources\cite{martorana2022aligning}. Nevertheless, the automatic enrichment of metadata when only the metadata is available is a challenging task because much of the contextual information is missing, and the underlying data cannot be used to help find the most appropriate annotations. Large Language Models (LLMs) could be a helpful solution in this by leveraging their training data as background knowledge. Moreover, Retrieval Augmented Generation (RAG) systems could further integrate external knowledge - for example from knowledge graphs, controlled vocabularies and glossaries - to enhance the LLMs flexibility across different domains. 

In this year SemTab challenge, participants in the ``Metadata to KG'' track aim to annotate tables using only table metadata (e.g. column and table names) without accessing the underlying data. This approach tests the ability to enrich metadata effectively under similar conditions imposed by restricted access data. To guide our investigation we have formulated the following research questions:

\begin{itemize}
    \item How do traditional semantic similarity methods compare to newer methods using Large Language Models (LLMs) in the semantic annotation of table metadata when the underlying data is not available?

    \item How does the temperature setting of LLMs impact their performance in this task? 
    
    \item How do different combinations of metadata information in traditional methods affect their performance? 
    
    \item How does the nature of the input data and glossary influence the results?
\end{itemize}

In the pages that follow, we further describe the importance of metadata, especially in settings where the underlying data is not available. We also introduce the term ``Column Vocabulary Association'', before discussing our main methodology and results.

\subsection{Dataless tables}
Recently we have seen more and more solutions for sharing data that is considered confidential or with restricted access. For example, there have been multiple developments of online Open Government Data (OGD) portals- e.g. Central Bureau for Statistics Netherlands (CBS)\footnote{\url{https://www.cbs.nl}}, U.S. Government's Open Data\footnote{\url{https://data.gov}} and the Canada's Open Government Portal\footnote{\url{https://search.open.canada.ca/data/}} portals aimed at enhancing innovation and research, by allowing users to investigate population data. Through OGD portals, the general public and researchers are able to use this data in a variety of fields, such as journalism, software development and research \citep{begany2021open}. The data in these portals is usually aggregated statistics, yet a significant amount of population data remains inaccessible to the public due to confidentiality concerns. This includes patient data, individual-level statistical data, and other type of data that contains Personable Identifiable Information (PII). 

Solutions have been proposed to facilitate the reuse of restricted access data. For instance, the Personal Health Train \cite{deist2020distributed} allows users to send their algorithms to where the data is stored. In this way the users do not have to store the data in personal devices, and also do not have direct access to it at any point of the analysis. However, users still need to know that the data exists and its structural details. Therefore, comprehensive metadata descriptions have a key role to facilitate this process. In previous research, we have introduced a metadata schema - the DataSet Variable Ontology (DSV) \cite{martorana2023advancing} - that captures information both at the dataset and variable levels, demonstrating that high-quality metadata can enable the discovery of restricted access data without compromising confidentiality. This approach involves annotating non-confidential information, such as column descriptions, the structure of the dataset and summary statistics.

The goal of this year SemTab challenge and the ``Metadata to KG'' track align with this research, as creating annotations is often challenging, requiring domain expertise and difficult to automate.

\subsection{Column Vocabulary Association (CVA)}
\label{column-vocabulary-association}
In the domain of Semantic Table Interpretation (STI), there are some well known challenged, including the Column Type Annotation (CTA), the Column Entity Annotation (CEA), and the Column Property Annotation (CPA) tasks. The \textbf{CTA} task involves identifying the semantic type (e.g. dates or geographical locations) of each column in the table. The \textbf{CEA} task, instead, involves linking each cell to an entity in a knowledge graph: for example, the cell containing the string ``New York'' to be linked to the WikiData entity for New York City (Q60). \textbf{CPA} requires the identification of relationships between columns of a table: for example, recognising that the columns with headers ``Mayor's Name'' and ``City'' are related to each other by the property eg:isMayorOf. 

In this work we introduce a new term: the Column Vocabulary Association (CVA) task. This task differs significantly from the previous ones because it does not rely on any information from the underlying data within the table. Instead, it aims to associate column headers with entries in controlled vocabularies purely based on semantic similarities. The distinction between the word \textit{association} and \textit{annotation} is also important in this context. Annotation typically refers to the labelling of data with tags or categories. In contrast, with the term association we refer more on the conceptual linkage between the textual information in a column header, and an external knowledge repository. This approach emphasizes understanding and leveraging the semantic meaning of the column headers themselves, without using any underlying data. By focusing on semantic similarities, we aim to create a method for interpreting and integrating restricted access datasets, to facilitate metadata enrichment and data discovery.

\section{SemTab Challenge}
Since its inception in 2019, the Semantic Web Challenge on Tabular Data to Knowledge Graph Matching, known as SemTab, has become a competitive event focused on benchmarking systems and approaches that support and enhance Semantic Table Interpretation (STI). The SemTab challenge typically consists of two main tracks: the \textit{``Accuracy Track''} and the \textit{``Dataset Track''}. 

The Accuracy Track places participants in a specific context with a predefined set of input data. Participants submit solutions for table annotation tasks, including Column Type Annotation (CTA), Column Entity Annotation (CEA), and Column Property Annotation (CPA). Previous editions of the SemTab challenge have featured interesting solutions. For example,  MTab uses a voting algorithm combines with probability models to improve annotations\cite{nguyen2019mtab}. Similarly, TorchicTab enables the annotation of tables with diverse structures through the use of an external knowledge graph\cite{dasoulas2023torchictab}. Another example is SemTex, which utilizes a hybrid annotation approach that improves performances, by analyzing relationships between entities in knowledge graphs and integrating them with gradient boosting analysis\cite{henriksen2023semtex}.

The Dataset Track, on the other hand, involves the submission of new datasets and benchmarks for evaluating various tasks related to the SemTab challenge and tabular data. In previous years, participants have produced versatile datasets applicable across multiple domains. One such example is the ToughTables Dataset, which consists of high-quality manually-curated tables with non-obviously linkable cells, such as those with ambiguous names, typos, and misspelled entity names\cite{cutrona2020tough}. There have also been domain-specific datasets. For example, the BiodivTab benchmark includes 50 datasets based on real-world biodiversity research data, enhanced by manual annotation\cite{abdelmageed2021biodivtab}. Another example is the TSOTSATable Dataset, that contains annotations of tables made using the FoodOn Ontology\cite{dooley2018foodon}, Open Research Knowledge Graph (ORKG)\cite{auer2020improving}, and Wikidata\cite{jiomekong2023semantic}.

This year, the SemTab challenge introduced a new track: the \textit{``Metadata to KG''} track. Participants of this task were asked to map table metadata to KGs without having access to the underlying data. This presents a unique challenge due to the limited available context, making traditional STI methods less applicable, as they typically rely on actual data for annotation. To better define this metadata-only task, we introduce the term \textbf{Column Vocabulary Association (CVA)}. As further described in section \ref{column-vocabulary-association},CVA involves annotating columns using solely KGs table metadata, without utilizing the underlying data. This approach is particularly relevant in scenarios where the data is confidential and cannot be accessed. 

The Metadata to KG track was divided into two rounds. Although the end goal remained the same (i.e. to annotate column headers with the appropriate term from KGs), each round features different input data and KGs. Below, we provide an explanatory summary of the two rounds. 

\subsection{Metadata to KG - Round 1}
Round 1 of the ``Metadata to KG'' track required participants to map a set of table metadata to DBpedia properties. Participants were provided with tables metadata and DBpedia properties files in both JSONL and OWL formats, all of which were accessible in the following GitHub repository\footnote{\url{https://github.com/sem-tab-challenge/2024/blob/main/data/metadata2kg/round1/README.md}}. 

The tables metadata file included information about 141 columns derived from different tables. For each column, the provided information included the column ID, column label, table ID, table name, and a list of the other column labels within the same table. The DBpedia properties file contained 2,881 properties. For each DBpedia property, the information included the property ID (the actual URI of the property in DBpedia), the property label, and the description. Below, we report examples of a table metadata entry (Listing \ref{lst:example-column-metadata}) and of DBpedia property (Listing \ref{lst:example-dbpedia-property}).

\vspace{10pt}
\begin{lstlisting}[caption={Example of table metadata},captionpos=b, label={lst:example-column-metadata}]
{
    "id": "58891288_0_1117541047012405958_Director(s)", 
    "label": "Director(s)", 
    "table_id": "58891288_0_1117541047012405958", 
    "table_name": "Film", 
    "table_columns": ["Rank", "Title", "Year", "Director(s)", "Overall Rank"]
}
\end{lstlisting}
\vspace{10pt}

\begin{lstlisting}[caption={Example of DBpedia property},captionpos=b, label={lst:example-dbpedia-property}]
{
    "id": "http://dbpedia.org/ontology/director", 
    "label": "film director", 
    "desc": "A film director is a person who directs the making of a film."
}
\end{lstlisting}

Additionally, the SemTab organizers supplied a sample table metadata file and a sample ground truth for validation and testing purposes, and a Python evaluation script. The objective was to develop approaches for mapping each table metadata with up to 5 DBpedia properties for each column, based on semantic similarities and relevance, and then rank the mappings from the most to least accurate. The evaluation script assessed the mappings by calculating two metrics: hit@1, which checks if the first mapping is correct, and hit@5, which checks if the correct mapping is within the top five. Participants tested their systems on sample metadata (containing only 9 columns) and submitted their complete results to the track organisers for evaluation against the overall ground truth.

\subsection{Metadata to KG - Round 2}
Round 2 of the ``Metadata to KG'' track introduced a level of complexity by using a collection of custom vocabularies for the mapping task. Participants were again provided with tables metadata and custom vocabularies files in JSONL and OWL formats, accessible in the following GitHub repository\footnote{\url{https://github.com/sem-tab-challenge/2024/blob/main/data/metadata2kg/round2/README.md}}. 

In this round, the tables metadata file contained 1181 entries (one entry corresponds to one column) from various datasets, with each column having the same information as in round 1 included: column ID, column label, table ID, table name, and the other columns labels. The custom vocabularies file consisted of 1192 entries, where each entry had, again, the same information as in round 1: ID (in this case not a URI, but a minted ID), label and description. The tables metadata included a very diverse set of topics: including but not limited to: COVID-19 clinical trials, Indian movies ratings and Saudia Arabia stock exchange data. As in Round 1, the SemTab organizers supplied a sample table metadata file and a sample ground truth for validation and testing purposes, alongside a Python evaluation script. The objective was to map each table column metadata to up to 5 relevant custom vocabulary terms based on semantic similarities and relevance, and rank these mappings by accuracy. The evaluation script again used hit@1 and hit@5 metrics to assess the quality of the mappings.

\section{Methods}

In this section, we outline our methodology, which considers the CVA task as a textual information retrieval challenge. Given that most of the table metadata (including label, table name, and table columns) and glossary information (label and description) are described in text, the goal is to retrieve the most similar glossary entries from the glossary file based on the table metadata.

Our approach involves two main methods: one utilizing LLMs (both open and commercial) and another employing a traditional semantic similarity method using SentenceBERT. We experimented with three GPT models (\texttt{gpt-3.5-turbo-0125} - \texttt{gpt-4o} - \texttt{gpt-4-turbo}), two LLama models (\texttt{llama3-70b} - \texttt{llama3-8b}), a Gemma model (\texttt{gemma-7b}) and a Mixtral model (\texttt{mixtral-8x7b}). Additionally, we varied the temperature settings (0.5, 0.75, 1.0, 1.25, and 1.5) for the LLMs to examine the impact of creativity on task performance. 


All our tests were conducted in a complete zero-shot setting, with no pretraining of the models and no examples provided through assistant instructions or user prompts. This approach is a key feature of our method. We chose this strategy because we aim to develop a method that is not domain-specific. Pretraining models with specific examples from particular mappings and vocabularies could bias the reported accuracy towards that specific domain. Instead, we aim to propose a method that can be applied more in general settings, regardless of the data domain or vocabulary.

\subsection{CVA with LLMs}
\textbf{Prompt Engineering}\\
Through trial and error, we developed effective prompts, both user queries and assistant instructions. We found that repeating some information from the assistant instructions within the prompt resulted in more precise results by ensuring the models only used the data we provided, thus minimizing hallucinations. Both the prompt and instructions specified to return the 5 most similar glossary entries for each metadata. Below, we show the instructions given to the assistants and the query template for the user prompt used in Round 1 of the SemTab Challenge. The instructions and prompts for Round 2, which are quite similar, can be found on the GitHub page. \\

\vspace{10pt}
\begin{tcolorbox}[height=0.55\textheight, left=2mm, right=2mm, fontupper=\ttfamily\scriptsize]
\textbf{ASSISTANT INSTRUCTIONS} \\
Your task is to match column metadata to DBpedia properties.\\
The full set of DBpedia properties will be provided in the vector. \\
Columns metadata, instead, will be provided by the user and it will contain the following information: column ID, column label, table ID, table name and the labels of the other columns within that table.
The matching between the column and the DBpedia properties is to be made based on the semantic similarities between the metadata (i.e. what the column express), and DBpedia properties. \\
You can add multiple properties, but no more 5. \\
Return the results in the following format: \\
'colID': '00000\_0\_0000\_XXX', 'propID': ['http://dbpedia.org/ontology/PROPERTY\_ID', ..., 'http://dbpedia.org/ontology/PROPERTY\_ID']. \\
Sort the matched DBpedia in descending order of relevance, starting with the most relevant. \\
Choose ONLY from the DBpedia properties. \\
Return ONLY the results, no other text. \\
Return results for each and every single column metadata. \\
\\
\\
\textbf{QUERY TEMPLATE} \\
Based on the instruction given to you, find the most relevant DBpedia property, for each of the following metadata in json format: \\
\{input\_metadata\} \\
Each json element is an independent column metadata. The metadata do not have any relationship, so the matching with the DBpedia properties should only be based on the information provided within its own metadata. \\
You can add multiple properties, but no more 5. \\
Return the results in the following format: \\
'colID': '00000\_0\_0000\_XXX', 'propID': ['http://dbpedia.org/ontology/PROPERTY\_ID', ..., 'http://dbpedia.org/ontology/PROPERTY\_ID']. \\
Sort the matched DBpedia in descending order of relevance, starting with the most relevant. \\
Choose ONLY from the DBpedia properties provided in the vector. \\
Return ONLY the results, no other text. \\
Return results for each and every single column metadata. \\

\end{tcolorbox}
\vspace{10pt}

\textbf{Model-Temperature Selection} \\
We ran the queries three times for each LLM and temperature combination, then evaluated the preliminary performance using an evaluation script and groundtruth provided by the organisers. Based on these results, we selected the best-performing LLM-temperature combination to compute results on the full dataset. \\

\textbf{CVA on Full Metadata Set} \\
In the first round, we input the complete glossary JSON file containing the DBpedia properties into the vector. We then processed 25 metadata entries at a time when using the OpenAI API, while for the open-source LLMs, each metadata entry was added individually. This approach was taken to minimize the cost of running some of the more expensive OpenAI models (4o and 4turbo). In the second round, given the larger size of the glossary, we split it into smaller, topic-based glossaries. We created 75 smaller glossary files and divided the full metadata set into 75 corresponding files. Each metadata file was then processed one at a time against the vector containing the 75 glossary files. \\

\subsection{Semantic similarity using SentenceBERT} 
Our second method involved computing the semantic similarity between table metadata and glossary entries using SentenceBERT\cite{reimers2019sentence}. First, we generated a vector representation for each metadata and glossary entry. Next, we calculated the cosine similarity between the embedding of each table metadata and the glossary entries to identify the top 5 glossary entries with the highest cosine similarity scores.

The initial steps of this method posed two challenges: 1) determining which table metadata and glossary information to use for generating the vectors, and 2) deciding how to vectorize the textual information — whether to concatenate all textual content before vectorization or to vectorize each part separately and then sum the vectors to form the final embedding. To address these questions, we experimented with different combinations of textual information, computed the cosine similarity, and evaluated the results against the groundtruth using the evaluation script provided by the organizers. Based on these findings, we selected the best performing combinations to use across the full metadata set.

\subsection{Evaluation}

The evaluation was conducted using a script provided by the track organizers. This script computed the accuracy of the generated mappings for a sample metadata file and a sample ground truth. It calculated two metrics: hit@1 and hit@5. To reiterate, users are supposed to generate the 5 most relevant mapping between the table metadata and the glossary, sorted from the most relevant. Hit@1 checks if the first mapping (thus the one considered to be the most relevant) is correct, while hit@5 checks if the correct mapping is among the top five results. Participants were then asked to generate the mappings for the entire table metadata file and submit them to the organizers. The organizers can then run the evaluation script again using the complete ground truth, which has not yet been shared with the participants.

\section{Results}

In the following sections, we present the preliminary challenge's results. These results show the accuracy scores obtained from the evaluation script on the sample metadata and the sample groundtruth provided. At this point, we are not aware on how our methods performed for the full set of metadata file, as the complete groundtruth has not yet been provided to participants.

\subsection{CVA with LLMs}
Here we present the results from our initial analysis using different LLMs and temperature settings. We employed three models from OpenAI and four open-source models, testing them at five different temperatures, as detailed in the table below\ref{tab:llms-results}. The table shows the average accuracy results for each model-temperature combination, evaluated using the evaluation script with the sample metadata and sample ground truth. Each query was run three times per model-temperature combination, and accuracy results were then averages. The numbers in bold correspond to the best-performing model-temperature combinations. \texttt{gpt-4o} outperformed other models in both Rounds 1 and 2, specifically at temperatures 0.5, 0.75, and 1.0. We observed that the LLMs did not perform very well in Round 2. In the discussion section\ref{discussion} we explore possible reasons for this outcome. 
Based on these preliminary results, for Round 1, we used \texttt{gpt-4o} at temperatures 0.5, 0.75, and 1.0 on the full metadata file for final analysis. For Round 2, we used \texttt{gpt-4o} only at temperatures 0.5 and 0.75.

\begin{table}[h!]
    \caption{Results of different models for sample data in Round 1 and 2. The cells with the ``X'' refers to tries where the LLM could not compute the task, and either the API was not returning any results over a long period of time or, in the case of \texttt{gemma-7b} the model was returning ``failure'' message. In bold, instead, we show the best performing results. In the table h1 and h5 refers to Hit@1 and Hit@5 metrics from the evaluation script.}
    \begin{tabular}{ccccccccccc}\toprule
    \multirow{3}{*}{\begin{tabular}[c]{@{}l@{}}LLM\\ Models\end{tabular}} &
      \multicolumn{10}{c}{Round 1} \\ \cline{2-11} 
     &
      \multicolumn{2}{c}{0.5} &
      \multicolumn{2}{c}{0.75} &
      \multicolumn{2}{c}{1.0} &
      \multicolumn{2}{c}{1.25} &
      \multicolumn{2}{c}{1.5} \\
      \cmidrule(lr){2-3}\cmidrule(lr){4-5}\cmidrule(lr){6-7}\cmidrule(lr){8-9}\cmidrule(lr){10-11}
     &
      h1 &
      h5 &
      h1 &
      h5 &
      h1 &
      h5 &
      h1 &
      h5 &
      h1 &
      h5 \\ \midrule
gpt-3.5-turbo-0125  & 0.42 & 0.42 & 0.33 & 0.42 & 0.47 & 0.53 & 0.41 & 0.41 & 0.36 & 0.36 \\
gpt-4o              & \textbf{0.59} & \textbf{0.89} & \textbf{0.62} & \textbf{0.81} & \textbf{0.64} & \textbf{0.75} & 0.47 & 0.67 & 0.11 & 0.22 \\
gpt-4-turbo         & 0.58 & 0.58 & 0.56 & 0.56 & 0.61 & 0.61 & 0.59 & 0.59 & 0.36 & 0.36 \\
llama3-8b           & 0.22 & 0.44 & 0.22 & 0.44 & 0.22 & 0.44 & 0.33 & 0.44 & 0.33 & 0.33 \\
llama3-70b          & 0.11 & 0.11 & 0.11 & 0.22 & 0.11 & 0.22 & 0.11 & 0.22 & 0.11 & 0.11 \\
gemma-7b            & 0.37 & 0.4  & 0.44 & 0.52 & 0.33 & 0.47 & 0.48 & 0.61 & 0.37 & 0.53 \\
mixtral-8x7b        & 0.33 & 0.33 & 0.44 & 0.56 & 0.33 & 0.44 & 0.44 & 0.44 & 0.44 & 0.44 \\
\hhline{===========}
    \multirow{3}{*}{} &
      \multicolumn{10}{c}{Round 2} \\ \cline{2-11} 
     &
      \multicolumn{2}{c}{0.5} &
      \multicolumn{2}{c}{0.75} &
      \multicolumn{2}{c}{1.0} &
      \multicolumn{2}{c}{1.25} &
      \multicolumn{2}{c}{1.5} \\
      \cmidrule(lr){2-3}\cmidrule(lr){4-5}\cmidrule(lr){6-7}\cmidrule(lr){8-9}\cmidrule(lr){10-11}
     &
      h1 &
      h5 &
      h1 &
      h5 &
      h1 &
      h5 &
      h1 &
      h5 &
      h1 &
      h5 \\ \midrule
gpt-3.5-turbo-0125  & 0.67 & 0.67 & 0.3  & 0.33 & 0.63 & 0.67 & 0.67 & 0.67 & X & X \\
gpt-4o              & \textbf{0.73} & \textbf{1}    & \textbf{0.45} & \textbf{1}    & 0.45 & 0.58 & 0.54 & 0.64 & X & X \\
gpt-4-turbo         & 0    & 0    & X    & X    & X    & X    & X    & X    & X & X \\
llama3-8b           & 0    & 0    & 0    & 0    & 0    & 0    & 0    & 0    & 0 & 0 \\
llama3-70b          & 0    & 0    & 0    & 0    & 0    & 0    & 0    & 0    & 0 & 0 \\
gemma-7b            & X    & X    & X    & X    & X    & X    & X    & X    & X & X \\
mixtral-8x7b        & 0    & 0    & 0    & 0    & 0    & 0    & 0    & 0    & 0 & 0 \\
\bottomrule
    \end{tabular}
    \label{tab:llms-results}
\end{table}

\subsection{CVA with SentenceBERT}

Below we show the results from our initial analysis with SentenceBERT. Table \ref{tab:embeddings-results} includes the possible combinations of information from the table metadata and the glossary, and the accuracy results for both Round 1 and 2, which we obtained by running the evaluation script against the ground truth for the sample metadata file. We used these results to find the best performing combinations, which were then applied to the full metadata file.

In Round 1, we did not have a single combination that performed best for both hit@1 and hit@5. The best hit@1 (0.56) is obtained when we use the column label and/or table name to represent the table metadata embedding, and use the property label for the DBpedia property embedding. The best hit@5 (0.67) is obtained when we use the sum of the vectors of the column label and the table name as the table metadata embedding, and use the vector of the property label as the DBpedia property embedding.

For Round 2, the best hit@1 and hit@5 are both obtained when we use the sum of the vectors of the column label and the table name as the table metadata embedding, and encode the vocabulary description as the vocabulary embedding. Based on this results, we did perform SentenceBERT on the full data for Round 1. For Round 2, instead, we sent the results from SentenceBERT for final analysis using the setting with the sum of the vectors of the column label and the table name as the table metadata embeddings.

\begin{table}[h!]
    \caption{Results of different embedding combination for sample data in Round 1 and 2. In bold we are showing the best performing results. In the table h1 and h5 refers to Hit@1 and Hit@5 metrics from the evaluation script.}
    \begin{tabular}{cccccc}\toprule
        \multicolumn{1}{c}{\multirow{2}{*}{\begin{tabular}[c]{@{}c@{}}Metadata\\ Embeddings\end{tabular}}} & 
        \multicolumn{1}{c}{\multirow{2}{*}{\begin{tabular}[c]{@{}c@{}}Glossary\\ Embeddings\end{tabular}}} &
        \multicolumn{2}{c}{Round 1} & 
        \multicolumn{2}{c}{Round 2}
        \\\cmidrule(lr){3-4}\cmidrule(lr){5-6}
\multicolumn{1}{c}{} & \multicolumn{1}{c}{} & h1  & h5  & h1 & h5\\\midrule
encode(label)                       & encode(label)                & \textbf{0.56} & 0.56 & 0.36 & 0.55 \\
encode(label)                       & encode(lable + desc)         & 0.22 & 0.56 & 0.45 & 0.82 \\
encode(label + table\_name)         & encode(label)                & \textbf{0.56} & 0.56 & 0.09 & 0.27 \\
encode(label + table\_name)         & encode(lable + desc)         & 0.33 & 0.44 & 0.64 & 0.73 \\
encode(label)                       & encode(desc)                 & 0.11 & 0.33 & 0.45 & 0.82 \\
encode(label + table\_name)         & encode(desc)                 & 0.22 & 0.44 & 0.64 & 0.73 \\
encode(label) + encode(table\_name) & encode(desc)                 & 0 & 0.33 & \textbf{0.64} & \textbf{0.91} \\
encode(label) + encode(table\_name) & encode(desc) + encode(label) & 0.22 & 0.44 & 0.55 & 0.91 \\
encode(label) + encode(table\_name) & encode(label)                & 0.44 & \textbf{0.67} & 0.27 & 0.45 \\\bottomrule
    \end{tabular}
    \label{tab:embeddings-results}
\end{table}

\section{Discussion}
\label{discussion}
During our analysis, we came across several interesting points that we would like to report in this section. Firstly, regarding the LLM prompt engineering, we found that repeating some of the same sentences in both the assistant instructions and user prompts improved the LLM's ability to follow instructions more precisely and avoid hallucinations. Specifically, this approach helped prevent the addition of entries not included in the glossaries.

Additionally, we observed notable differences in the type of data included between the Round 1 and Round 2 of the challenge. In round 1, there was no clear link between the table metadata file and the glossary. The glossary consisted of DBpedia properties, while the table metadata appeared to be a random collection from various data sources. In round 2, instead, there was a very clear connection between the table metadata and the glossary. It seemed that the table metadata was designed to match the glossary or vice versa, likely because the data was collected from institutions/organizations/repositories that build their own glossaries to describe their data. These differences had an effect in the performances of the methods proposed in this work. LLMs, particularly \texttt{gpt-4o}, performed much better in round 1, where it was important to leverage the LLM's background knowledge to find the most relevant mappings. In round 2, however, the semantic similarities method was sufficient and sometimes even outperformed LLMs. This was due to the high degree of semantic similarity between the column headers and the glossary, as both probably originated from the same institution and were intentionally made to be similar.

Lastly, we want to highlight differences related to the temperature settings we used for the LLMs. We found that temperatures below 1 performed better than higher temperatures. Temperature regulates the LLM's creativity, where 0 represents no creativity and 2 indicates full creativity. In many cases, and particularly with \texttt{gpt-4-turbo} and \texttt{gemma-7b}, a temperature of 1.5 resulted in no outputs or error messages. This suggests that lower temperatures may lead to better performance for these types of tasks, although further investigation is needed.

\section{Conclusion}
In this study, we investigated different methods for mapping column headers to glossaries when the underlying data is unavailable. Our approach operates in a zero-shot setting, meaning we do not pretrain models or provide examples of correct mappings. This allows us to evaluate model performance across different domains more broadly. We introduce the concept of ``Column Vocabulary Association'' (CVA) and distinguish it from other STI tasks. Additionally, we analyze how different temperatures of LLMs and the types of input data and glossaries impact the performance in the CVA task.

Our findings suggest that LLMs perform well when there are no clear connections between the data and glossaries used for mapping (as observed in Round 1), leveraging their extensive background knowledge. Instead, for CVA tasks where the input metadata and glossaries are closely related (as in Round 2), traditional semantic similarity methods (e.g. SentenceBERT) may perform better than LLMs. Preliminary results indicate that open source models still lag behind commercial ones, with less accurate performance compared to the GPT models. In future work we aim to further evaluate LLM performance by calculating additional metrics introduced in \cite{martorana2024text} to assess the consistency of the LLM performances.

\begin{acknowledgments}
We acknowledge that ChatGPT was utilized to generate and debug part of the python and latex code utilised in this work. This work is funded by the Netherlands Organisation of Scientific Research (NWO), ODISSEI Roadmap project: 184.035.014. 
\end{acknowledgments}

\bibliography{bibliography.bib}

\begin{thebibliography}{19}
\expandafter\ifx\csname natexlab\endcsname\relax\def\natexlab#1{#1}\fi
\providecommand{\url}[1]{\texttt{#1}}
\providecommand{\href}[2]{#2}
\providecommand{\path}[1]{#1}
\providecommand{\DOIprefix}{doi:}
\providecommand{\ArXivprefix}{arXiv:}
\providecommand{\URLprefix}{URL: }
\providecommand{\Pubmedprefix}{pmid:}
\providecommand{\doi}[1]{\href{http://dx.doi.org/#1}{\path{#1}}}
\providecommand{\Pubmed}[1]{\href{pmid:#1}{\path{#1}}}
\providecommand{\bibinfo}[2]{#2}
\ifx\xfnm\relax \def\xfnm[#1]{\unskip,\space#1}\fi
\bibitem[{Shwartz-Ziv and Armon(2022)}]{shwartz2022tabular}
\bibinfo{author}{R.~Shwartz-Ziv}, \bibinfo{author}{A.~Armon},
\newblock \bibinfo{title}{Tabular data: Deep learning is not all you need},
\newblock \bibinfo{journal}{Information Fusion} \bibinfo{volume}{81} (\bibinfo{year}{2022}) \bibinfo{pages}{84--90}.
\bibitem[{Boeckhout et~al.(2018)Boeckhout, Zielhuis, and Bredenoord}]{boeckhout2018fair}
\bibinfo{author}{M.~Boeckhout}, \bibinfo{author}{G.~A. Zielhuis}, \bibinfo{author}{A.~L. Bredenoord},
\newblock \bibinfo{title}{The fair guiding principles for data stewardship: fair enough?},
\newblock \bibinfo{journal}{European journal of human genetics} \bibinfo{volume}{26} (\bibinfo{year}{2018}) \bibinfo{pages}{931--936}.
\bibitem[{Mons(2018)}]{mons2018data}
\bibinfo{author}{B.~Mons}, \bibinfo{title}{Data stewardship for open science: Implementing FAIR principles}, \bibinfo{publisher}{Chapman and Hall/CRC}, \bibinfo{year}{2018}.
\bibitem[{Lamprecht et~al.(2020)Lamprecht, Garcia, Kuzak, Martinez, Arcila, Martin Del~Pico, Dominguez Del~Angel, Van De~Sandt, Ison, Martinez et~al.}]{lamprecht2020towards}
\bibinfo{author}{A.-L. Lamprecht}, \bibinfo{author}{L.~Garcia}, \bibinfo{author}{M.~Kuzak}, \bibinfo{author}{C.~Martinez}, \bibinfo{author}{R.~Arcila}, \bibinfo{author}{E.~Martin Del~Pico}, \bibinfo{author}{V.~Dominguez Del~Angel}, \bibinfo{author}{S.~Van De~Sandt}, \bibinfo{author}{J.~Ison}, \bibinfo{author}{P.~A. Martinez}, et~al.,
\newblock \bibinfo{title}{Towards fair principles for research software},
\newblock \bibinfo{journal}{Data Science} \bibinfo{volume}{3} (\bibinfo{year}{2020}) \bibinfo{pages}{37--59}.
\bibitem[{Wilkinson et~al.(2016)Wilkinson, Dumontier, Aalbersberg, Appleton, Axton, Baak, Blomberg, Boiten, da~Silva~Santos, Bourne et~al.}]{wilkinson2016fair}
\bibinfo{author}{M.~D. Wilkinson}, \bibinfo{author}{M.~Dumontier}, \bibinfo{author}{I.~J. Aalbersberg}, \bibinfo{author}{G.~Appleton}, \bibinfo{author}{M.~Axton}, \bibinfo{author}{A.~Baak}, \bibinfo{author}{N.~Blomberg}, \bibinfo{author}{J.-W. Boiten}, \bibinfo{author}{L.~B. da~Silva~Santos}, \bibinfo{author}{P.~E. Bourne}, et~al.,
\newblock \bibinfo{title}{The fair guiding principles for scientific data management and stewardship},
\newblock \bibinfo{journal}{Scientific data} \bibinfo{volume}{3} (\bibinfo{year}{2016}) \bibinfo{pages}{1--9}.
\bibitem[{Martorana et~al.(2022)Martorana, Kuhn, Siebes, and van Ossenbruggen}]{martorana2022aligning}
\bibinfo{author}{M.~Martorana}, \bibinfo{author}{T.~Kuhn}, \bibinfo{author}{R.~Siebes}, \bibinfo{author}{J.~van Ossenbruggen},
\newblock \bibinfo{title}{Aligning restricted access data with fair: a systematic review},
\newblock \bibinfo{journal}{PeerJ Computer Science} \bibinfo{volume}{8} (\bibinfo{year}{2022}) \bibinfo{pages}{e1038}.
\bibitem[{Begany et~al.(2021)Begany, Martin, and Yuan}]{begany2021open}
\bibinfo{author}{G.~M. Begany}, \bibinfo{author}{E.~G. Martin}, \bibinfo{author}{X.~J. Yuan},
\newblock \bibinfo{title}{Open government data portals: Predictors of site engagement among early users of health data ny},
\newblock \bibinfo{journal}{Government Information Quarterly} \bibinfo{volume}{38} (\bibinfo{year}{2021}) \bibinfo{pages}{101614}.
\bibitem[{Deist et~al.(2020)Deist, Dankers, Ojha, Marshall, Janssen, Faivre-Finn, Masciocchi, Valentini, Wang, Chen et~al.}]{deist2020distributed}
\bibinfo{author}{T.~M. Deist}, \bibinfo{author}{F.~J. Dankers}, \bibinfo{author}{P.~Ojha}, \bibinfo{author}{M.~S. Marshall}, \bibinfo{author}{T.~Janssen}, \bibinfo{author}{C.~Faivre-Finn}, \bibinfo{author}{C.~Masciocchi}, \bibinfo{author}{V.~Valentini}, \bibinfo{author}{J.~Wang}, \bibinfo{author}{J.~Chen}, et~al.,
\newblock \bibinfo{title}{Distributed learning on 20 000+ lung cancer patients--the personal health train},
\newblock \bibinfo{journal}{Radiotherapy and Oncology} \bibinfo{volume}{144} (\bibinfo{year}{2020}) \bibinfo{pages}{189--200}.
\bibitem[{Martorana et~al.(2023)Martorana, Kuhn, Siebes, and Van~Ossenbruggen}]{martorana2023advancing}
\bibinfo{author}{M.~Martorana}, \bibinfo{author}{T.~Kuhn}, \bibinfo{author}{R.~Siebes}, \bibinfo{author}{J.~Van~Ossenbruggen},
\newblock \bibinfo{title}{Advancing data sharing and reusability for restricted access data on the web: introducing the dataset-variable ontology},
\newblock in: \bibinfo{booktitle}{Proceedings of the 12th Knowledge Capture Conference 2023}, \bibinfo{year}{2023}, pp. \bibinfo{pages}{83--91}.
\bibitem[{Nguyen et~al.(2019)Nguyen, Kertkeidkachorn, Ichise, and Takeda}]{nguyen2019mtab}
\bibinfo{author}{P.~Nguyen}, \bibinfo{author}{N.~Kertkeidkachorn}, \bibinfo{author}{R.~Ichise}, \bibinfo{author}{H.~Takeda},
\newblock \bibinfo{title}{Mtab: Matching tabular data to knowledge graph using probability models},
\newblock \bibinfo{journal}{arXiv preprint arXiv:1910.00246}  (\bibinfo{year}{2019}).
\bibitem[{Dasoulas et~al.(2023)Dasoulas, Yang, Duan, and Dimou}]{dasoulas2023torchictab}
\bibinfo{author}{I.~Dasoulas}, \bibinfo{author}{D.~Yang}, \bibinfo{author}{X.~Duan}, \bibinfo{author}{A.~Dimou},
\newblock \bibinfo{title}{Torchictab: Semantic table annotation with wikidata and language models},
\newblock in: \bibinfo{booktitle}{CEUR Workshop Proceedings}, \bibinfo{organization}{CEUR Workshop Proceedings}, \bibinfo{year}{2023}, pp. \bibinfo{pages}{21--37}.
\bibitem[{Henriksen et~al.(2023)Henriksen, Khorsid, Nielsen, St{\"u}ck, S{\o}rensen, and Pelgrin}]{henriksen2023semtex}
\bibinfo{author}{E.~G. Henriksen}, \bibinfo{author}{A.~M. Khorsid}, \bibinfo{author}{E.~Nielsen}, \bibinfo{author}{A.~M. St{\"u}ck}, \bibinfo{author}{A.~S. S{\o}rensen}, \bibinfo{author}{O.~Pelgrin},
\newblock \bibinfo{title}{Semtex: A hybrid approach for semantic table interpretation.},
\newblock in: \bibinfo{booktitle}{SemTab@ ISWC}, \bibinfo{year}{2023}, pp. \bibinfo{pages}{38--49}.
\bibitem[{Cutrona et~al.(2020)Cutrona, Bianchi, Jim{\'e}nez-Ruiz, and Palmonari}]{cutrona2020tough}
\bibinfo{author}{V.~Cutrona}, \bibinfo{author}{F.~Bianchi}, \bibinfo{author}{E.~Jim{\'e}nez-Ruiz}, \bibinfo{author}{M.~Palmonari},
\newblock \bibinfo{title}{Tough tables: Carefully evaluating entity linking for tabular data},
\newblock in: \bibinfo{booktitle}{International Semantic Web Conference}, \bibinfo{organization}{Springer}, \bibinfo{year}{2020}, pp. \bibinfo{pages}{328--343}.
\bibitem[{Abdelmageed et~al.(2021)Abdelmageed, Schindler, and K{\"o}nig-Ries}]{abdelmageed2021biodivtab}
\bibinfo{author}{N.~Abdelmageed}, \bibinfo{author}{S.~Schindler}, \bibinfo{author}{B.~K{\"o}nig-Ries},
\newblock \bibinfo{title}{Biodivtab: A table annotation benchmark based on biodiversity research data.},
\newblock in: \bibinfo{booktitle}{SemTab@ ISWC}, \bibinfo{year}{2021}, pp. \bibinfo{pages}{13--18}.
\bibitem[{Dooley et~al.(2018)Dooley, Griffiths, Gosal, Buttigieg, Hoehndorf, Lange, Schriml, Brinkman, and Hsiao}]{dooley2018foodon}
\bibinfo{author}{D.~M. Dooley}, \bibinfo{author}{E.~J. Griffiths}, \bibinfo{author}{G.~S. Gosal}, \bibinfo{author}{P.~L. Buttigieg}, \bibinfo{author}{R.~Hoehndorf}, \bibinfo{author}{M.~C. Lange}, \bibinfo{author}{L.~M. Schriml}, \bibinfo{author}{F.~S. Brinkman}, \bibinfo{author}{W.~W. Hsiao},
\newblock \bibinfo{title}{Foodon: a harmonized food ontology to increase global food traceability, quality control and data integration},
\newblock \bibinfo{journal}{npj Science of Food} \bibinfo{volume}{2} (\bibinfo{year}{2018}) \bibinfo{pages}{23}.
\bibitem[{Auer et~al.(2020)Auer, Oelen, Haris, Stocker, D’Souza, Farfar, Vogt, Prinz, Wiens, and Jaradeh}]{auer2020improving}
\bibinfo{author}{S.~Auer}, \bibinfo{author}{A.~Oelen}, \bibinfo{author}{M.~Haris}, \bibinfo{author}{M.~Stocker}, \bibinfo{author}{J.~D’Souza}, \bibinfo{author}{K.~E. Farfar}, \bibinfo{author}{L.~Vogt}, \bibinfo{author}{M.~Prinz}, \bibinfo{author}{V.~Wiens}, \bibinfo{author}{M.~Y. Jaradeh},
\newblock \bibinfo{title}{Improving access to scientific literature with knowledge graphs},
\newblock \bibinfo{journal}{Bibliothek Forschung und Praxis} \bibinfo{volume}{44} (\bibinfo{year}{2020}) \bibinfo{pages}{516--529}.
\bibitem[{Jiomekong et~al.(2023)Jiomekong, Melie, Tapamo, and Camara}]{jiomekong2023semantic}
\bibinfo{author}{A.~Jiomekong}, \bibinfo{author}{U.~Melie}, \bibinfo{author}{H.~Tapamo}, \bibinfo{author}{G.~Camara},
\newblock \bibinfo{title}{Semantic annotation of tsotsatable dataset.},
\newblock in: \bibinfo{booktitle}{SemTab@ ISWC}, \bibinfo{year}{2023}, pp. \bibinfo{pages}{15--20}.
\bibitem[{Reimers and Gurevych(2019)}]{reimers2019sentence}
\bibinfo{author}{N.~Reimers}, \bibinfo{author}{I.~Gurevych},
\newblock \bibinfo{title}{Sentence-bert: Sentence embeddings using siamese bert-networks},
\newblock \bibinfo{journal}{arXiv preprint arXiv:1908.10084}  (\bibinfo{year}{2019}).
\bibitem[{Martorana et~al.(2024)Martorana, Kuhn, Stork, and van Ossenbruggen}]{martorana2024text}
\bibinfo{author}{M.~Martorana}, \bibinfo{author}{T.~Kuhn}, \bibinfo{author}{L.~Stork}, \bibinfo{author}{J.~van Ossenbruggen},
\newblock \bibinfo{title}{Text classification of column headers with a controlled vocabulary: leveraging llms for metadata enrichment},
\newblock \bibinfo{journal}{arXiv preprint arXiv:2403.00884}  (\bibinfo{year}{2024}).

\end{thebibliography}

\end{document}